\documentclass{egpubl}
\usepackage{sca2026}

\usepackage[T1]{fontenc}
\usepackage{dfadobe}
\usepackage{amsmath,amssymb}
\usepackage{booktabs}
\usepackage{capt-of}
\usepackage{balance}

\SpecialIssuePaper

\CGFccby

\makeatletter
\renewcommand\p@copyrightTextLong{}
\renewcommand\p@copyrightTextShort{}
\renewcommand\p@copyrightTextShortEven{}
\makeatother

\biberVersion
\BibtexOrBiblatex

\usepackage[
  backend=biber,
  bibstyle=EG,
  citestyle=alphabetic,
  backref=true
]{biblatex}

\electronicVersion
\PrintedOrElectronic

\usepackage{graphicx}
\ifpdf
\fi

\usepackage{egweblnk}
\usepackage[capitalise,noabbrev]{cleveref}

\title[Multi-View Face and Gesture Animation]%
      {Multi-View Face and Gesture Animation with
Dynamic Gaussians}

\author[Javanmardi et al.]
{\parbox{\textwidth}{\centering
Alireza Javanmardi$^{1\dagger}$\orcid{0009-0008-4926-1566},
Vippin Kumar Jeetmal$^{2\dagger}$\orcid{0009-0002-6270-2975},
Christen Millerdurai$^{1}$\orcid{0009-0001-1653-8126},
Alain Pagani$^{1}$\orcid{0000-0002-5136-0837}
and Didier Stricker$^{1,2}$\orcid{0009-0004-8794-6858}
}
\\
{\parbox{\textwidth}{\centering
$^1$German Research Center for Artificial Intelligence (DFKI), $^2$RPTU \\
$^\dagger$Equal contribution
}
}
}

\begin{document}

\teaser{%
  \centering
  \includegraphics[width=\textwidth]{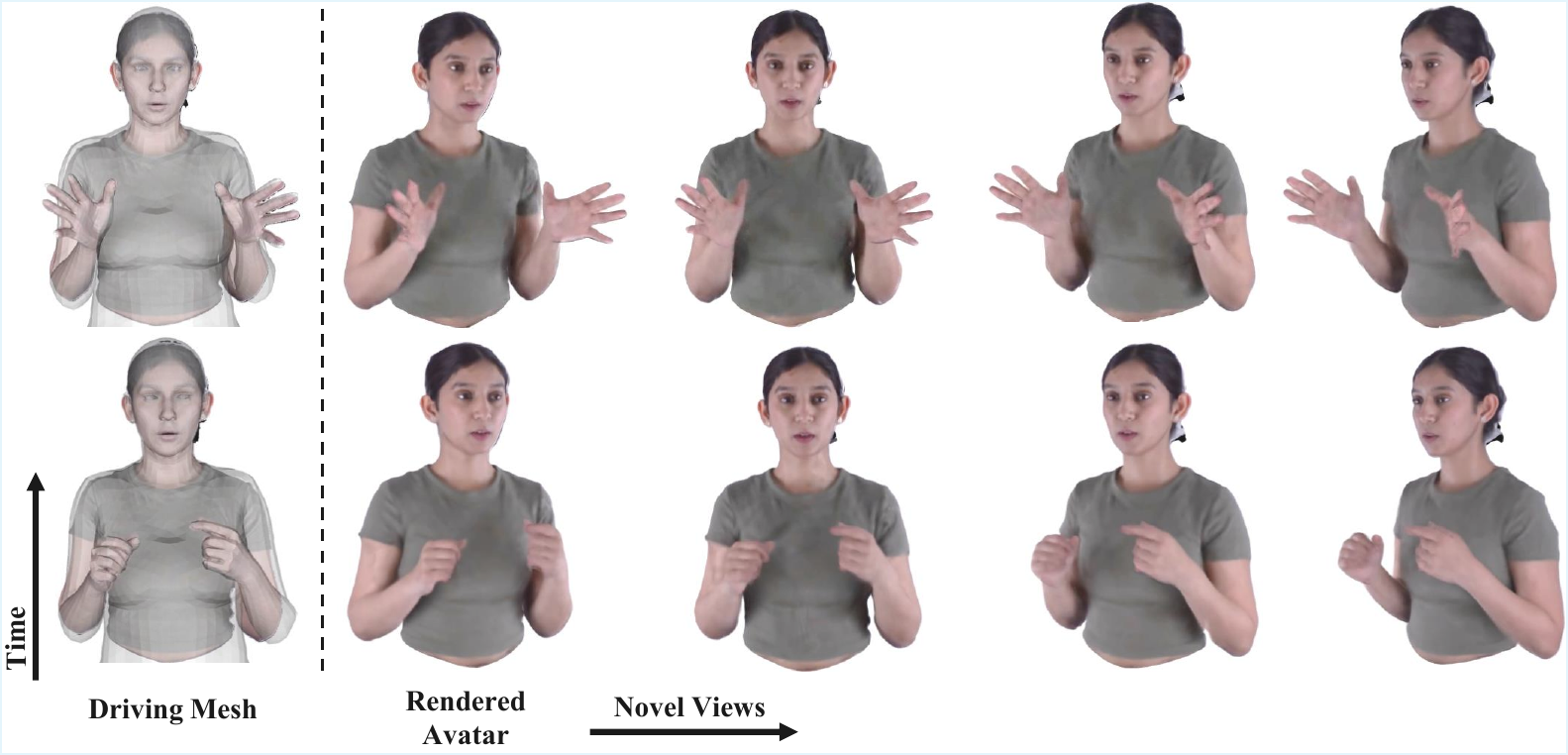}
  \captionof{figure}{\textbf{Multi-view face and gesture animation.}
  Given a time-varying driving mesh (left), our method (\textbf{MVFGA})
  animates a reconstructed 3D avatar learned from multi-view data and
  synthesizes geometry-consistent novel-view renderings (right),
  faithfully reproducing facial expressions and hand gestures.}
  \label{fig:teaser}
}

\maketitle
\begin{abstract}
Creating photorealistic 3D human avatars with realistic upper-body motion remains challenging. 
Existing approaches either focus on the head and overlook hand gestures, or reconstruct the full body but fail to preserve fine-grained facial fidelity and hand pose accuracy. 
As a result, current methods struggle to capture the subtle dynamics of facial expressions and hand gestures that are crucial for natural human communication. 
While methods based on full-body parametric models enable avatar reconstruction from monocular or multi-view inputs, they often lack accurate facial animation and detailed hand articulation.
To address these limitations, we propose MVFGA, a novel multi-view-consistent pipeline for generating realistic upper-body avatars. 
Our approach models the face and hands separately and fuses them with a parametric upper-body mesh model, enabling the capture of fine-grained facial expressions and hand poses for accurate upper-body avatar reconstruction.
We then splat 3D Gaussians onto the obtained mesh, enabling high-quality rendering of dynamic avatars from novel viewpoints.
Furthermore, we introduce MVFGA-MoCap, a multi-view upper-body motion capture dataset featuring controlled facial expression sequences, diverse hand gestures, and free-form communication. 
Experiments show that MVFGA generates visually realistic avatars with high-fidelity facial expressions and hand motions, outperforming baselines for upper-body avatar animation. Project page: \url{https://dfki-av.github.io/MVFGA/}


\end{abstract}   
\section{Introduction}
Animating humans in photorealistic 3D has become an increasingly important research problem due to the growing demand for authentic and engaging digital interactions across applications such as virtual reality, gaming, remote communication, and content creation.
Beyond entertainment, realistic human avatars are becoming essential in high-stakes scenarios such as remote medical training, surgical teleoperation, virtual education, and collaborative design, where subtle facial expressions and precise hand gestures play a critical role in effective communication \cite{10445546, app15063290}. 
In these settings, digital avatars act as proxies for real individuals in virtual environments and are often personalized to closely resemble the target user. 
However, imperfect appearance or motion can reduce user comfort rather than enhance immersion, a response commonly attributed to the uncanny valley effect \cite{mori1970bukimi}, which describes negative reactions to near-human entities with subtle perceptual or behavioral flaws.


The main challenge is to disentangle the actor's appearance from their motion, including pose and facial expressions. 
This disentanglement enables animation at inference time using novel motions, either from new sequences of the same actor or transferred from a different actor.
Consequently, recent research has largely followed two directions. One line of work focuses on animating facial expressions only, producing talking-head or portrait-level animations~\cite{wang2021one,ma2024follow,kirschstein2024diffusionavatars}. 
Another line aims to generate holistic full-body animations~\cite{siarohin2021motion,wang2022latent,qiu2025LHM}. 
While these approaches have achieved impressive results, they often fall short in capturing realistic upper-body communication, which critically involves both expressive facial dynamics and precise hand gestures.
This limitation significantly reduces their applicability in scenarios where hand motion and facial expression are tightly coupled.

\begin{figure*}[!t]
  \centering
  \includegraphics[width=1.95\columnwidth]{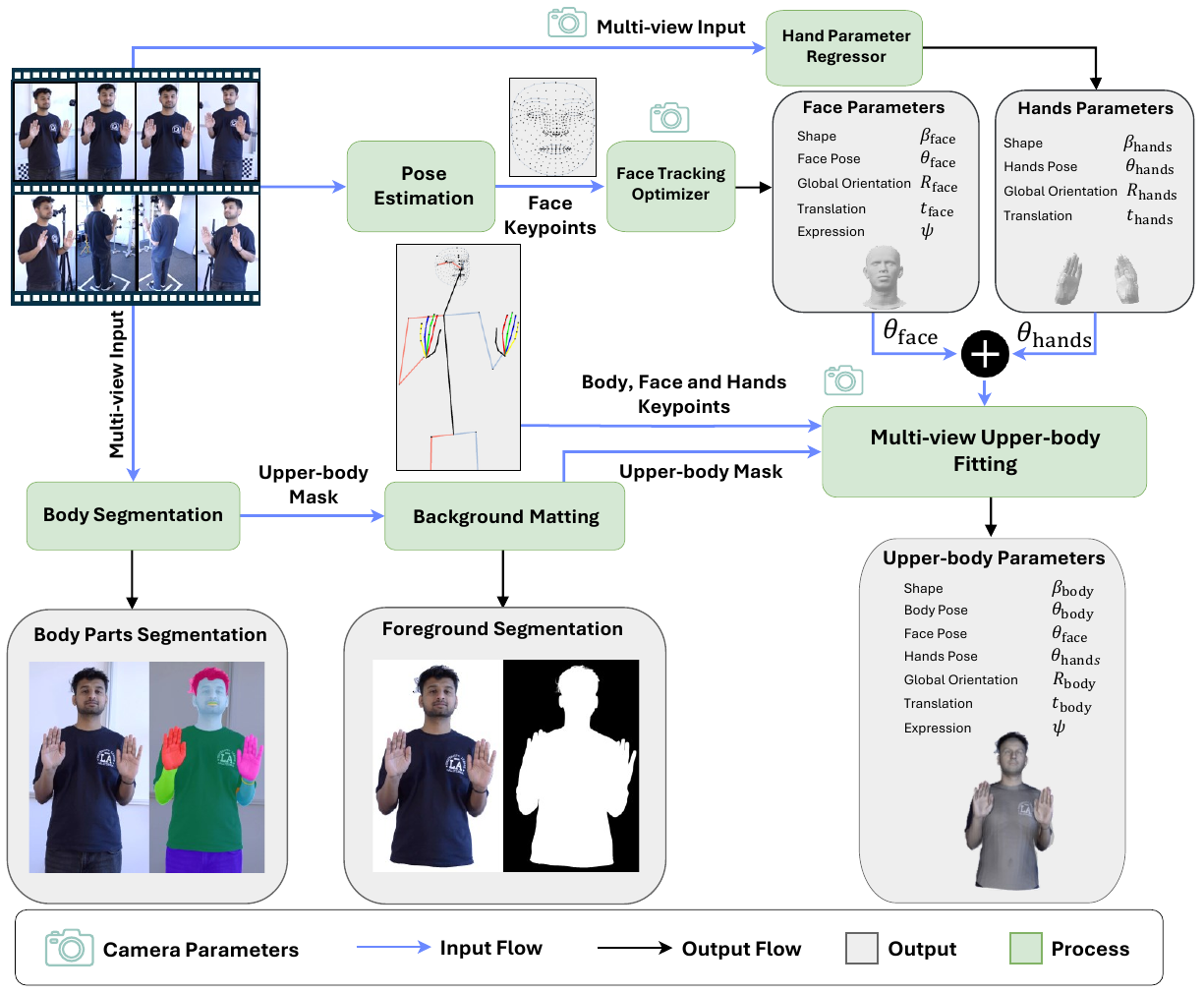}
\vspace{-5pt}
\caption{\textbf{Upper-body mesh generation pipeline}: We perform keypoint extraction, body-part segmentation, and background matting, followed by refined face and hand fitting. The resulting face, hand, and body parameters are fused to obtain a complete upper-body representation with accurate shape, pose, and global translation.}
  \label{Fig:pipe1}
\end{figure*}

Recent progress in human animation has been advanced by both graphics-based \cite{qian2024gaussianavatars,junkawitsch2025eva, Zhang_2025_ICCV} and generative approaches \cite{taubner2025cap4d, buehler2024cafca, tu2025stableanimator}. While these methods have shown promising performance in one-/few-shot settings, they still struggle to produce lifelike avatars with multi-view consistency. In practice, reconstructed avatars often suffer from degraded facial fidelity, incomplete finger articulation, and view-dependent artifacts, which become particularly noticeable when generative backbones inpaint unseen regions. These limitations highlight the need for a framework that can reconstruct and render reliable, high-quality avatars suitable for real-world applications. \\
To address these limitations, we introduce \textbf{Multi-View Face and Gesture Animation with Dynamic Gaussians (MVFGA)}, a multi-view-consistent framework for upper-body avatar reconstruction and reenactment as shown in \cref{fig:teaser}. 
Our end-to-end pipeline builds an animatable upper-body parametric model by integrating facial expressions and hand articulations into a unified representation, enabling detailed facial and hand motion.
We then splat 3D Gaussians~\cite{kerbl3Dgaussians} onto the mesh to synthesize photorealistic dynamic avatars under novel viewpoints. 
The generated avatars can be driven from a monocular video while preserving multi-view-consistent appearance, enabling real-time and immersive applications.
To support training and evaluation, we introduce \textbf{MVFGA-MoCap}, a comprehensive multi-view dataset for upper-body face and gesture animation. 
It includes synchronized videos of 15 subjects captured by 17 calibrated cameras from diverse viewpoints, covering controlled facial expressions, diverse hand gestures, and free-form communication.
Our main contributions are summarized as follows:
\begin{itemize}
    \item We propose an animatable upper-body parametric model that separately integrates face and hand parameterizations into a unified representation.
    \item We present a novel multi-view pipeline for reconstructing the upper-body motion and appearance of a human actor, enabling photorealistic upper-body avatar synthesis and novel-view rendering.    
    \item We introduce \textbf{MVFGA-MoCap}, a multi-view upper-body motion-capture dataset featuring controlled facial expressions and diverse hand gestures.
\end{itemize}

\section{Related Work}

\subsection{Graphics-based Avatar Animation}
Early approaches typically rely on parametric face or body models, such as FLAME \cite{FLAME:SiggraphAsia2017}, or SMPL \cite{SMPL:2015}, where personalized geometry is reconstructed and animated using pose and expression parameters \cite{Thies_2016_CVPR}. Once the geometry is reconstructed, appearance is modeled via texture mapping and rendered using rasterization-based pipelines. These methods offer strong structural consistency and real-time performance; however, their visual fidelity is often limited by the expressiveness of the rendering pipeline, making it challenging to capture fine-scale details such as hair, accessories, and subtle appearance variations.

Neural rendering techniques have significantly advanced realism by replacing explicit mesh-based representations with implicit neural representations. Neural Radiance Fields (NeRFs) \cite{mildenhall2020nerf} and their dynamic variants enable high-quality, view-consistent avatar reconstruction and animation \cite{Gafni_2021_CVPR, park2021hypernerf, athar2023flame}. While these approaches substantially improve visual quality compared to traditional graphics-based methods, their reliance on computationally expensive ray sampling results in slow rendering speeds, limiting their practicality for real-time applications and deployment-critical scenarios.

More recently, 3D Gaussian Splatting has emerged as an efficient alternative that combines high visual fidelity with fast rendering performance \cite{kerbl3Dgaussians}. Several works leverage Gaussian representations for head avatar reconstruction and animation, achieving improved efficiency over NeRF-based methods \cite{qian2024gaussianavatars, xu2023gaussianheadavatar, teotia2025audio, aneja2025scaffoldavatar}. Other approaches extend Gaussian splatting to full-body animation, producing coherent holistic renderings but often lacking fine-grained facial and hand details \cite{Pang_2024_CVPR, qiu2025LHM, moon2024exavatar}. Conversely, face-centric methods achieve high-quality facial rendering but neglect articulated hand gestures. Additionally, recent few-shot \cite{zielonka2025synshot} or single-view Gaussian-based approaches \cite{he2025lam, Zhang_2025_ICCV} require large-scale training datasets and often exhibit limited generalization to novel viewpoints.
In contrast, we bridge the gap between face-centric and full-body methods by enabling multi-view–consistent upper-body reenactment with expressive facial dynamics and articulated hand gestures.

\subsection{Generative Avatar Animation}

Generative-based avatar animation methods aim to synthesize animated faces or bodies directly from data, without explicit 3D reconstruction. Early approaches predominantly adopt adversarial learning frameworks, where motion is represented using 2D or 3D keypoints extracted from driving frames and used to warp source features for animation \cite{siarohin2019first, Javanmardi_2024_BMVC, wang2021one, siarohin2021motion}. While effective for talking-head synthesis, these methods often struggle with large pose variations and full-body motion due to the absence of explicit 3D structure.

More recent works explore latent-space manipulation via GAN inversion \cite{wang2022latent, 10645735} or leverage diffusion models to improve visual fidelity and controllability \cite{xu2024magicanimate, hu2024animate, zhu2024champ, tu2025stableanimator, mimicmotion2024}. Diffusion-based approaches benefit from strong image priors and support diverse conditioning signals such as pose, motion, and audio. However, they typically require substantial computational resources and slow sampling procedures, limiting real-time applicability. Moreover, achieving long-term temporal consistency remains challenging, and the resulting avatars are not explicitly animatable in 3D.
A recent line of work combines generative models with 3D representations such as Gaussian Splatting to enable few-shot 3D avatar animation \cite{taubner2025cap4d, kirschstein2025avat3r}. Despite promising results, these approaches often suffer from hallucination of unseen regions, hindering consistent 3D avatar creation and reliable retargeting. In contrast, we adopt a graphics-based, explicitly animatable representation for improved controllability and efficiency.
\begin{figure*}[!t]
  \centering
  \includegraphics[width=2.05\columnwidth]{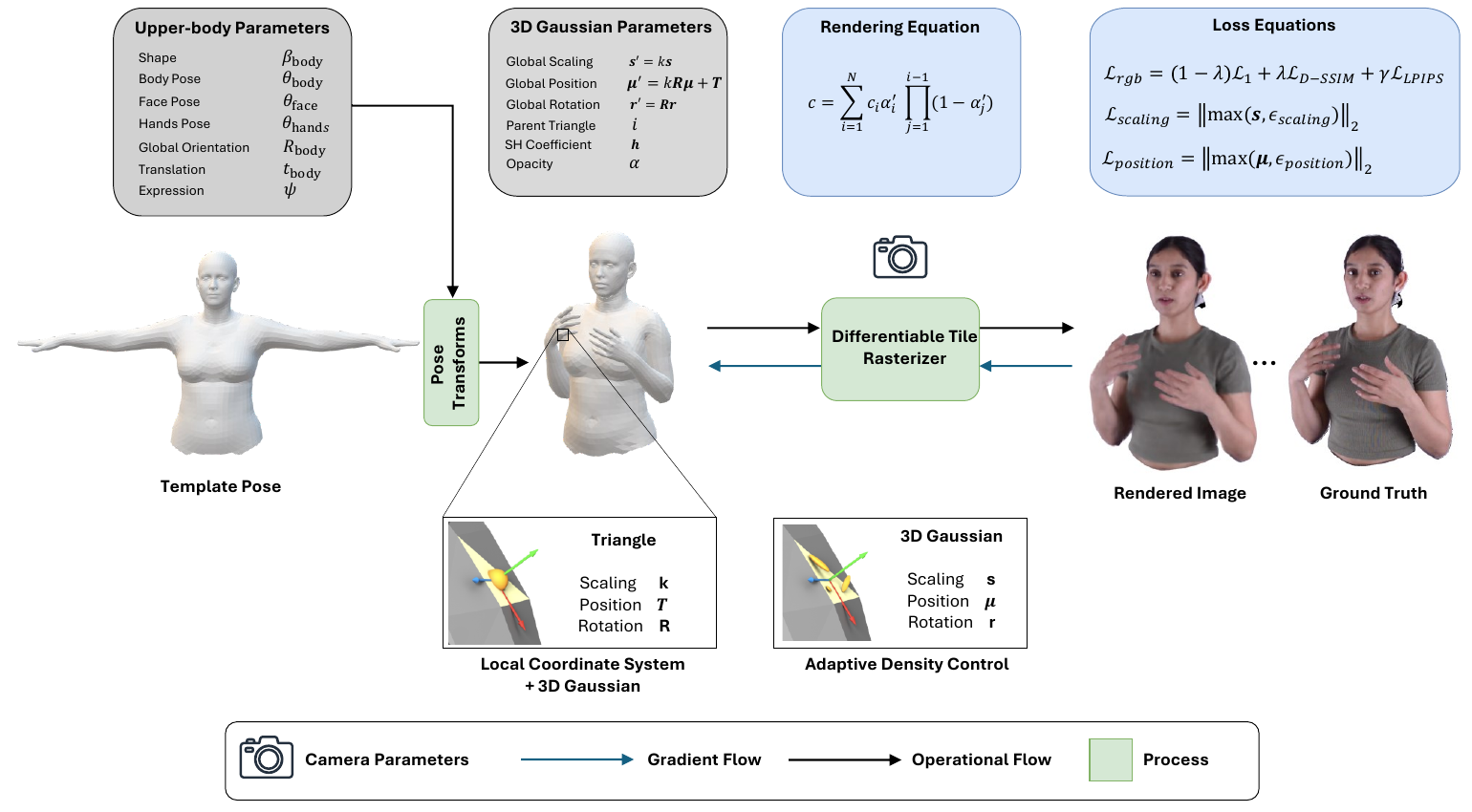}
\caption{\textbf{Overview of our avatar synthesis pipeline}. Given multi-view images and corresponding
upper-body parameters, following GaussianAvatar \cite{qian2024gaussianavatars} the template mesh is posed into the deformed space.
Each triangle is then assigned a 3D Gaussian representation. These Gaussians are
rasterized using a tile-based rasterizer to produce rendered images, which are
supervised using an RGB reconstruction loss. An adaptive density control mechanism
is employed during training to dynamically densify or prune the Gaussians
for efficient and accurate representation.}
  \label{fig:pipe2}
\end{figure*}

\begin{figure}[h]
  \centering
  \includegraphics[width=1\columnwidth]{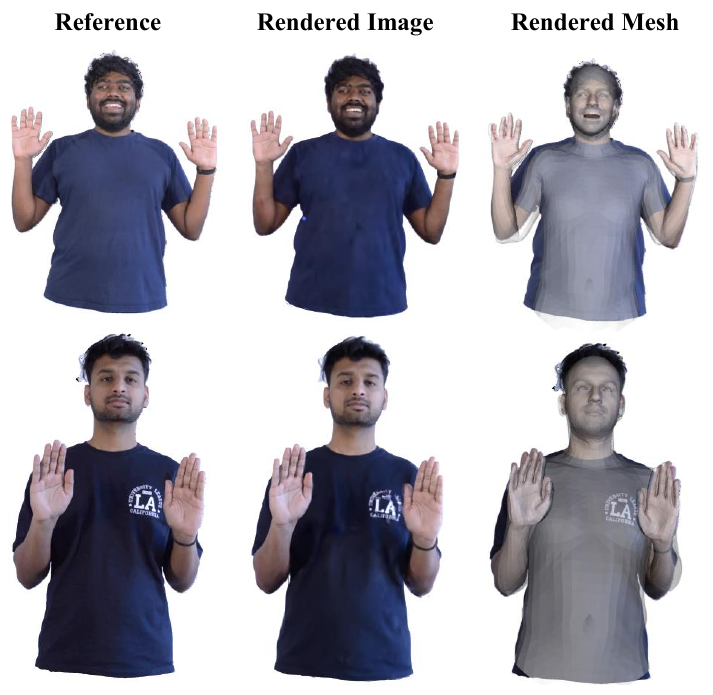}
\caption{\textbf{Upper-body mesh fitting}. From left to right: reference, rendered image, and rendered fitted mesh.}
  \label{fig:mesh}
\end{figure}

\section{Methodology}
\label{sec:methodology}
Our aim is to create and animate a realistic 3D upper-body avatar of an actor that maintains high fidelity under novel viewpoints.
First, we capture the actor using a synchronized multi-view rig and fit our upper-body parametric model to the recordings, yielding a temporally consistent mesh that deforms with the actor's motion (Sec.~\ref{subsec:preprocessing}).
Next, we place 3D Gaussians on the mesh surface and optimize them to reconstruct the actor's appearance across time and camera viewpoints (Sec.~\ref{subsec:animation_pipeline}). 
This process produces a deformable Gaussian-splat representation, which can be animated at inference time under novel target motions and viewpoints. 
Finally, we describe the dataset collected during this process in Sec.~\ref{subsec:dataset}.

\subsection{Upper-body Mesh Reconstruction}
\label{subsec:preprocessing}
As shown in \cref{Fig:pipe1}, we reconstruct a detailed 3D upper-body mesh from temporally aligned multi-view RGB video captured by 17 calibrated cameras. 
Our pipeline begins by extracting 2D full-body keypoints in all views to obtain a robust pose initialization and reduce ambiguity caused by self-occlusions.
To capture expressive motion, we further refine regions where whole-body parametric fitting is less accurate. 
For the face, we estimate facial shape and expression parameters from the video using an optimization-based face tracking procedure. 
For the hands, we process the left and right hand regions independently using a transformer-based estimator that regresses hand pose and shape parameters.
Specifically, we use MICA~\cite{MICA:ECCV2022} for face parameters and HaMeR~\cite{pavlakos2024reconstructing} to estimate MANO~\cite{MANO:SIGGRAPHASIA:2017} hand parameters. 
A key challenge is that these methods are designed and optimized for monocular inputs, and therefore do not directly account for our calibrated multi-view camera intrinsics and extrinsics, nor do they produce view-consistent parameter estimates.
To address this, we generate multiple candidate estimates and select the one with the lowest aggregated multi-view 2D reprojection error, while rejecting outliers.
Next, we define our upper-body parametric model, a novel extension of SMPL-X~\cite{SMPL-X:2019} that restricts the full-body mesh topology to the upper body by removing vertices and faces that lie outside the upper-body region.
We preserve its parameterization, including the shape space as well as pose and expression parameters, making our model backward compatible with SMPL-X.
We fit this upper-body model to the actor by jointly optimizing the body parameters using the multi-view 2D keypoints together with the estimated face and hand parameters, while explicitly accounting for the calibrated camera intrinsics and extrinsics (Multi-view Upper-body Fitting in \cref{Fig:pipe1}). 
This yields a multi-view-consistent upper-body mesh with highly articulated face and hand regions (see \cref{fig:mesh}).
Finally, we apply background matting and semantic body-part segmentation to isolate the actor and generate a clean upper-body mask. This mask is used to optimize the 3D Gaussians and provides a high-fidelity matte for the rendered avatar.
Additional details are provided in the supplementary material.

\subsection{3D Gaussian Splatting Preliminary}
\label{subsec:preliminary}

We build on 3D Gaussian Splatting (3DGS)~\cite{kerbl3Dgaussians}, which reconstructs a scene from multi-view images and calibrated cameras using anisotropic 3D Gaussians. Each splat is centered at a mean $\boldsymbol{\mu}$ and represented by a covariance matrix $\boldsymbol{\Sigma}$, defining the density
\begin{equation}
G(\mathbf{x})=\exp\!\left(-\frac{1}{2}(\mathbf{x}-\boldsymbol{\mu})^{\top}\boldsymbol{\Sigma}^{-1}(\mathbf{x}-\boldsymbol{\mu})\right).
\label{eq:gaussian_def}
\end{equation}

To ensure $\boldsymbol{\Sigma}$ remains positive semi-definite during optimization, 3DGS parameterizes each Gaussian as an oriented ellipsoid with rotation $\mathbf{R}$ and scaling $\mathbf{S}$:
\begin{equation}
\boldsymbol{\Sigma}=\mathbf{R}\mathbf{S}\mathbf{S}^{\top}\mathbf{R}^{\top}.
\label{eq:covariance_param}
\end{equation}
In practice, each ellipsoid is stored as $\boldsymbol{\mu}\in\mathbb{R}^{3}$, a scaling vector $\mathbf{s}\in\mathbb{R}^{3}$, and a quaternion $\mathbf{q}\in\mathbb{R}^{4}$ (we denote its rotation matrix by $\mathbf{r}\in\mathbb{R}^{3\times 3}$).
For rendering, Gaussians are projected to the image plane and composited in depth order using alpha blending:
\begin{equation}
\mathbf{C}=\sum_{i=1}^{N}\mathbf{c}_{i}\alpha'_{i}\prod_{j=1}^{i-1}\left(1-\alpha'_{j}\right),
\label{eq:alpha_blending}
\end{equation}
where $\mathbf{c}_i$ is modeled with low-order spherical harmonics and $\alpha'_i$ is obtained from the projected 2D Gaussian weighted by opacity.



\subsection{Upper-body Avatar Animation Pipeline}
\label{subsec:animation_pipeline}
Our animation pipeline, as illustrated in \cref{fig:pipe2}, shows the upper-body mesh augmented with a set of surface-attached 3D Gaussians. 
Given the posed mesh and calibrated cameras, we render the Gaussians via a differentiable rasterizer and optimize their parameters to match the observed multi-view appearance. 
To maintain both efficiency and fidelity, we employ adaptive density control throughout training.

\subsubsection{Initialization.}
Following GaussianAvatar~\cite{qian2024gaussianavatars}, we adapt their 3D Gaussian initialization strategy to place 3D Gaussians on our upper-body parametric mesh model.
Specifically, each triangle of the mesh is initialized with a single 3D Gaussian splat positioned at the triangle center. 
Each Gaussian is parameterized in the local coordinate system of its parent triangle by a mean location $\mu$, a rotation matrix $\mathbf{r}$, and a scaling vector $\mathbf{s}$. At initialization, $\mu$ is set to the local origin, $\mathbf{r}$ is the identity rotation, and $\mathbf{s}$ is set to a unit scale. During rendering, these local parameters are transformed into the global coordinate system using:
\begin{align}
\mathbf{r}' &= \mathbf{R}\mathbf{r}, \label{eq:rotation_transform} \\
\mu' &= k\mathbf{R}\mu + \mathbf{T}, \label{eq:mean_transform} \\
\mathbf{s}' &= k\mathbf{s}, \label{eq:scale_transform}
\end{align}
where $\mathbf{R}$ and $\mathbf{T}$ denote the global rotation and translation of the corresponding mesh triangle, respectively, and $k$ is a scalar scale factor that describes the local-to-global scaling of the triangle.

\subsubsection{Adaptive Density Control}
To capture high-frequency appearance details that are not explicitly modeled by the mesh geometry, we employ an adaptive density control mechanism based on \cite{kerbl3Dgaussians, qian2024gaussianavatars} that dynamically adds or removes splats based on view-space positional gradients and opacity statistics. 
When a Gaussian is split or cloned during densification, the newly created splats inherit the same parent triangle as the original, ensuring that they remain consistently attached to the mesh surface. 
This association is maintained by storing the index of the parent triangle for each Gaussian. 
In addition, pruning operations remove splats with persistently low opacity to improve computational efficiency. 
To avoid artifacts in frequently occluded regions (e.g., the eyes), we enforce that each mesh triangle always retains at least one associated Gaussian splat, even after pruning.

\subsubsection{Optimization Objectives.}
The 3D Gaussian splats are rendered into RGB images using a differentiable tile-based rasterizer and supervised using a combination of pixel-wise and perceptual losses. The RGB reconstruction loss is defined as:
\begin{equation}
\mathcal{L}_{\text{rgb}} = (1-\lambda)\mathcal{L}_{1} + \lambda\,\mathcal{L}_{\text{D-SSIM}} + \gamma\,\mathcal{L}_{\text{LPIPS}},
\end{equation}
where $\lambda$ and $\gamma$ are loss weighting terms, $\mathcal{L}_{1}$ denotes the $\ell_{1}$ loss, $\mathcal{L}_{\text{D-SSIM}}$ is the differentiable SSIM loss, and $\mathcal{L}_{\text{LPIPS}}$ is the perceptual loss as suggested in~\cite{hu2024expressive}.

In addition, we incorporate regularization terms to ensure stable training. A position regularization loss constrains Gaussian means to remain close to their parent triangle:
\begin{equation}
\mathcal{L}_{\text{position}} = \left\lVert \max(\mu, \epsilon_{\text{position}}) \right\rVert_{2},
\end{equation}
where $\epsilon_{\text{position}} = 1$ allows small deviations due to triangle scaling. We further introduce a scaling regularization loss to prevent splats from becoming excessively large relative to their parent triangle:
\begin{equation}
\mathcal{L}_{\text{scaling}} = \left\lVert \max(\mathbf{s}, \epsilon_{\text{scaling}}) \right\rVert_{2},
\end{equation}
where $\epsilon_{\text{scaling}} = 0.6$ disables this penalty for sufficiently small splats.
The final objective is:
\begin{equation}
\mathcal{L} = \mathcal{L}_{\text{rgb}} + \lambda_{\text{position}}\mathcal{L}_{\text{position}} + \lambda_{\text{scaling}}\mathcal{L}_{\text{scaling}} .
\end{equation}

Here, $\lambda_{\text{position}}$ and $\lambda_{\text{scaling}}$ are weighting coefficients for the position and scaling regularizers, respectively. We apply these regularization terms only when the RGB reconstruction loss is active for the corresponding timestep.

\subsection{Dataset}
\label{subsec:dataset}
We collect a high-quality multi-view upper-body motion capture dataset targeting detailed facial expressions, hand articulation, and torso motion. 
Data are recorded in a calibrated 17-camera RGB studio setup, including 15 front-facing cameras spanning a $150^\circ$ arc and two rear cameras for improved body keypoint detection. 
All videos are synchronized and captured at $1920 \times 1080$ resolution and 25 FPS under uniform studio lighting. 
The dataset contains 15 participants (8 male, 7 female; ages 24--32) performing diverse facial expressions and two-handed gestures. 
Further details on capture, calibration, statistics, and post-processing are provided in the supplementary material.

\section{Experiments and Results}
\label{sec:experiments}

\subsection{Experimental Setup}
\label{subsec:experiment_setup}

We evaluate our framework on our multi-view upper-body dataset under three settings: (1) \textit{self-reenactment}, where we drive an avatar using a held-out sequence of the same subject with unseen poses and expressions and render the frontal view; (2) \textit{novel-view synthesis}, where we animate the avatar using motions from training sequences and render from a held-out camera viewpoint; and (3) \textit{cross-identity animation}, where we transfer poses and expressions from one subject to animate the avatar of another subject.
\subsection{Implementation Details}
\label{subsec:implementation_details}
Our pipeline begins by processing multi-view video to obtain upper-body representation. 
We extract the foreground subject using BiRefNet~\cite{zheng2024birefnet} and segment semantic body parts with Sapiens~\cite{khirodkar2024sapiens}. 
Initial 2D keypoints from MediaPipe~\cite{lugaresi2019mediapipe} are combined with detailed FLAME~\cite{FLAME:SiggraphAsia2017} face parameters (from MICA~\cite{MICA:ECCV2022}) and MANO~\cite{MANO:SIGGRAPHASIA:2017} hand parameters (from HaMeR~\cite{pavlakos2024reconstructing}) within the EasyMocap~\cite{dong2021fast} framework to produce temporally consistent SMPL-X ~\cite{SMPL-X:2019} parameters.
We then derive our upper-body parametric model by selecting the upper-body vertices using the SMPL-X part-segmentation map and discarding vertices and triangles outside the upper-body region.
During optimization of the Gaussians, we use the semantic part segmentation to mask out lower-body pixels in each view.
All parameters are optimized with Adam~\cite{kingma2017adam}, using the original learning rates for the Gaussians. 
The SMPL-X parameters are fine-tuned with component-specific learning rates: $1\times10^{-8}$ (global translation), $1\times10^{-4}$ (facial expressions), and $1\times10^{-6}$ (body pose). We set loss weights $\lambda=0.2$, $\gamma=0.04$, $\lambda_{\text{position}}=0.01$, and $\lambda_{\text{scaling}}=1$. 
Each avatar is optimized for 600K iterations on a single NVIDIA H100 GPU, taking approximately 8 hours.

\subsection{Baselines}
\label{subsec:baseline}
For comparison, we evaluate state-of-the-art methods for upper-body reenactment. We include recent generative approaches, AnimateAnyone~\cite{hu2024animate}, MagicAnimate~\cite{xu2024magicanimate}, and Champ~\cite{zhu2024champ}. We also compare against the graphics-based baseline GUAVA \cite{Zhang_2025_ICCV}, which explicitly supports upper-body avatar modeling. Most methods primarily target either face-only animation \cite{xu2023gaussianheadavatar, qian2024gaussianavatars} or full-body synthesis \cite{qiu2025LHM, lei2024gart} and do not perform reliably on our upper-body setting.

\begin{figure*}[h]
  \centering
  \includegraphics[width=2.09\columnwidth]{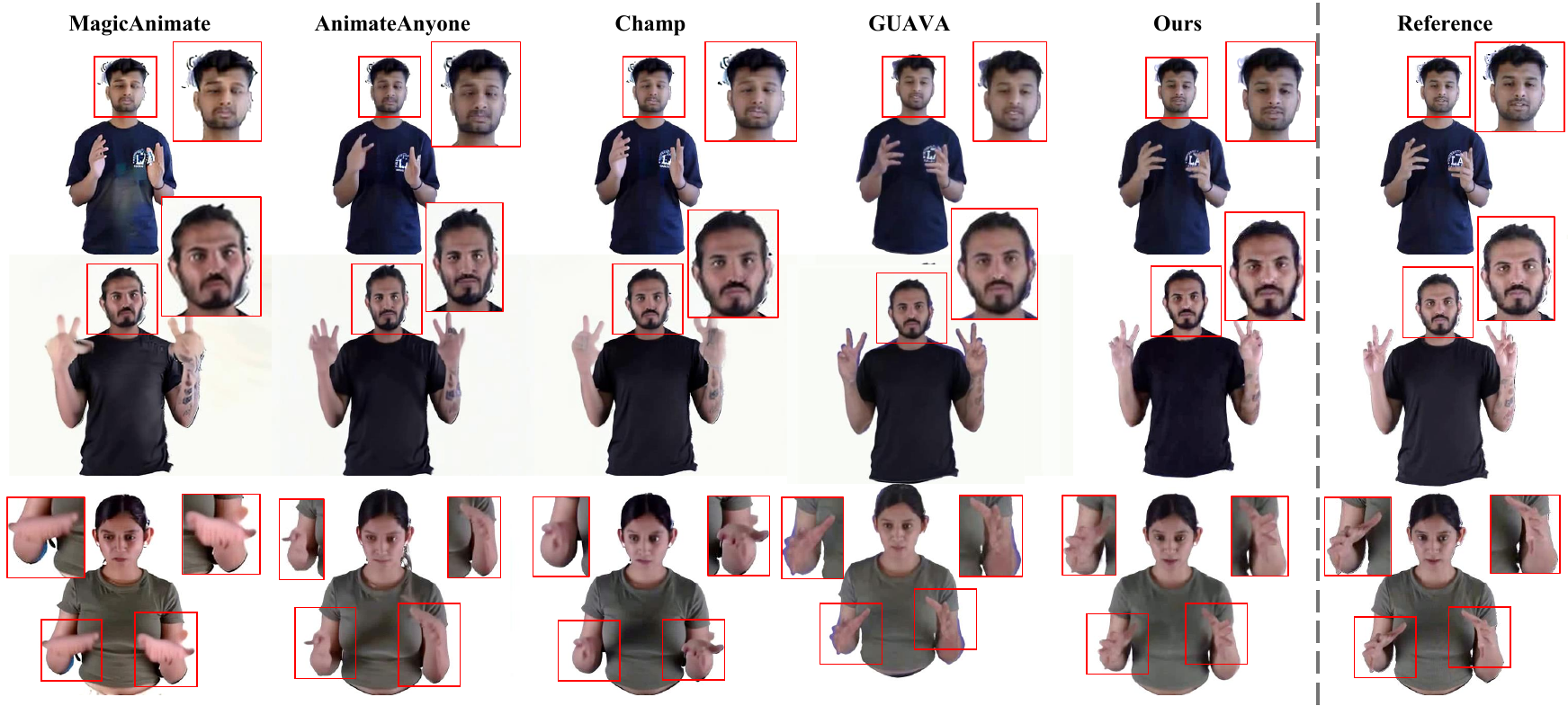}
\caption{\textbf{Qualitative comparison}. Our method captures facial expressions, hand gestures, and upper-body poses across diverse motions while better preserving appearance compared to MagicAnimate\cite{xu2024magicanimate}, AnimateAnyone \cite{hu2024animate}, Champ \cite{zhu2024champ}, and GUAVA \cite{Zhang_2025_ICCV}.}
  \label{fig:qualitative}
\end{figure*}
\begin{figure*}[!t]
  \centering
  \includegraphics[width=2.09\columnwidth]{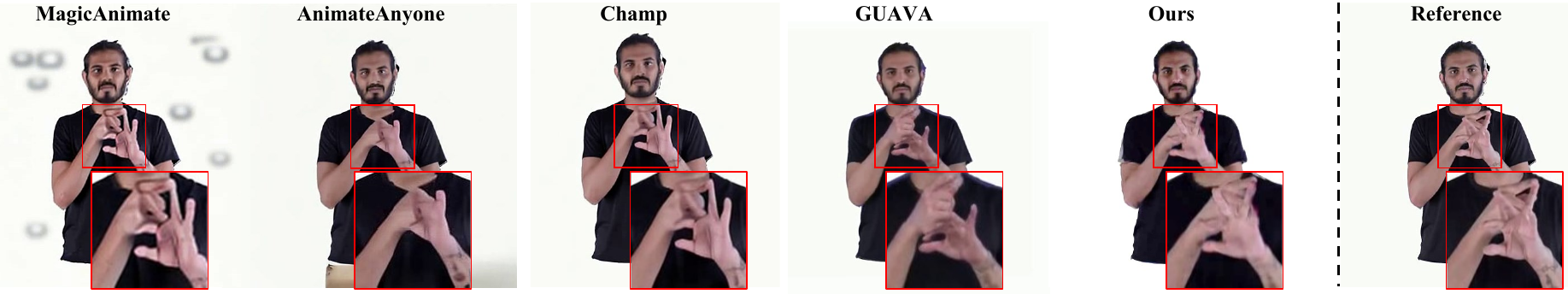}
\caption{\textbf{Qualitative comparison under complex hand articulation}. The highlighted region contains a challenging gesture with closely interacting fingers and strong self-occlusions. Compared with prior methods, our approach better preserves finger structure, hand pose accuracy, and overall appearance consistency.}
\vspace{-10pt}
\label{fig:handy}
\end{figure*}

\subsection{Metrics}
\label{subsec:Metrics}

We evaluate image quality using L1 error, Peak Signal-to-Noise Ratio (PSNR)~\cite{hore2010image}, Structural Similarity Index Measure (SSIM)~\cite{wang2004image}, and Learned Perceptual Image Patch Similarity (LPIPS). All metrics are computed using the Disco evaluation toolkit~\cite{wang2024disco}.
For pose accuracy, we report Average Keypoint Distance (AKD)~\cite{JMLR:v12:gashler11a}, computed on MediaPipe landmarks~\cite{lugaresi2019mediapipe} for the face, hands, and torso.
To assess facial identity preservation, we compute cosine similarity (CSIM) using ArcFace features~\cite{deng2019arcface}.
To evaluate temporal consistency, particularly for long-form animation rendering, we use Temporal Jittering Error (TJE)~\cite{javanmardi2026talkingpose}.

\begin{figure*}[!t]

  \centering
  \includegraphics[width=1.88\columnwidth]{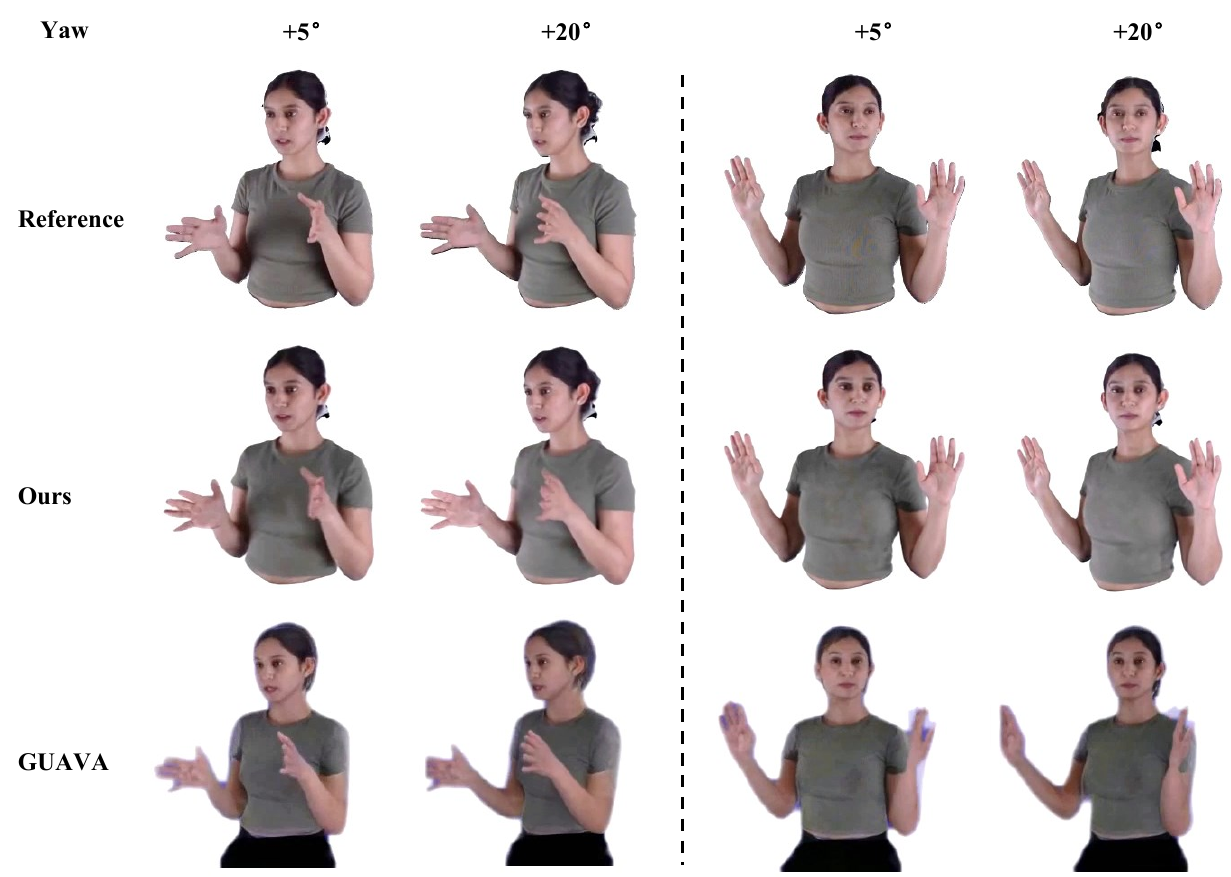}
\caption{\textbf{Novel-view synthesis.} Rendering at unseen yaw angles ($+5^\circ$, $+20^\circ$). Our method preserves geometry and appearance better than GUAVA 
\cite{Zhang_2025_ICCV}.}
\vspace{-18pt}
  \label{fig:novel}
\end{figure*}

\begin{figure}[!t]
  \centering
  \includegraphics[width=0.79\columnwidth]{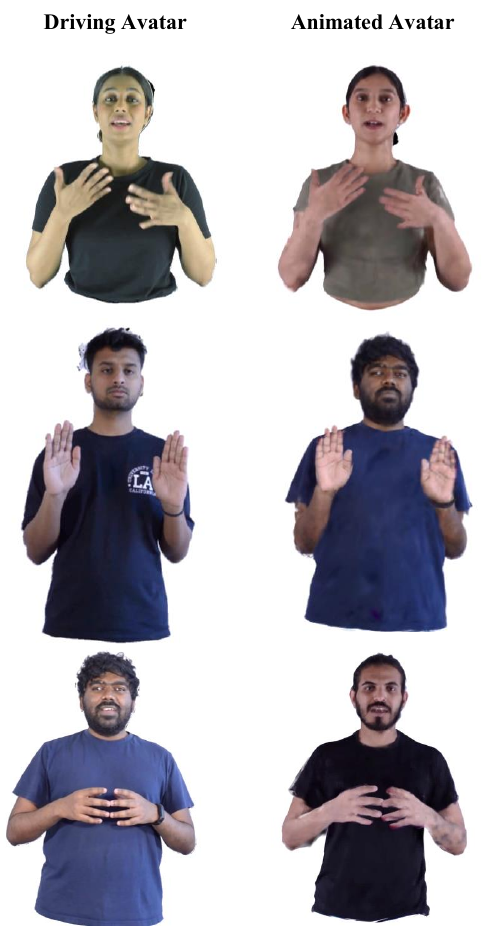}
\caption{\textbf{Cross-identity reenactment.} Motions from a driving subject are transferred to a target avatar while preserving identity and gestures.}
\vspace{-28pt}
\label{fig:cross}
\end{figure}

\begin{figure}[!t]

  \centering
  \includegraphics[width=0.9\columnwidth]{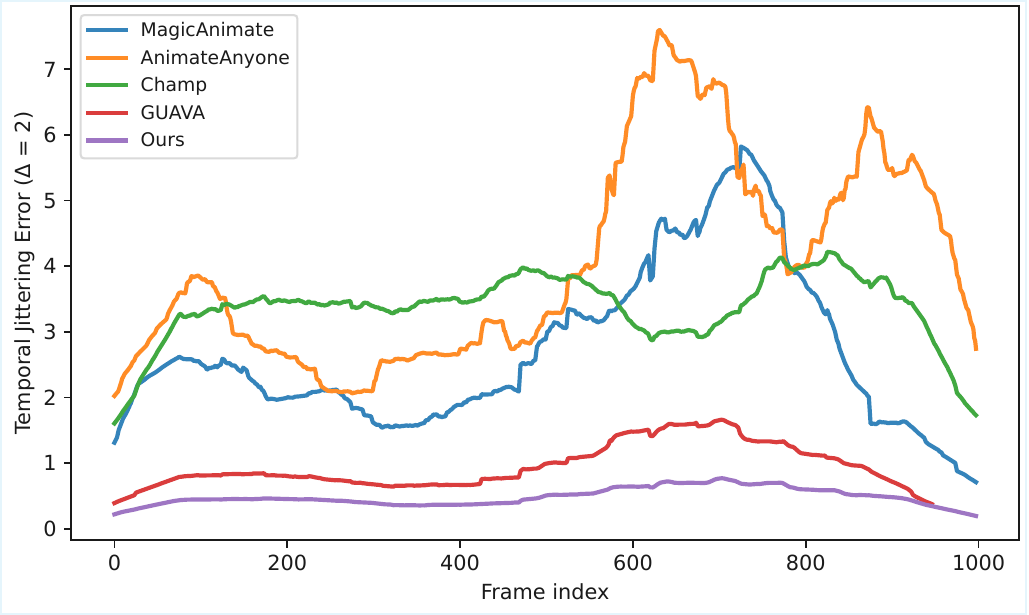}
\caption{\textbf{Temporal Jittering Error (TJE).} Lower values indicate better temporal consistency. TJE measures the discrepancy between motion differences in real and generated videos, capturing subtle jitter and flickering artifacts across frames.}
\vspace{-18pt}
  \label{fig:TJE}
\end{figure}

\subsection{Evaluation}
\label{subsec:Evaluation}

\noindent \textit{Protocol.}
For comparison methods, we use the first frame of each sequence as the source image and the remaining frames as the driving video, enabling frame-wise comparison across methods. As several generative baselines struggle with long video synthesis, we split each sequence into 100-frame clips for evaluation. We report results on four held-out identities.

\subsubsection{Quantitative Results}
\label{subsubsec:quantitative_results}
We report quantitative results using the metrics described in Sec.~\ref{subsec:Metrics} in \cref{tab:overall_comparison}. Our method achieves the best overall performance across all metrics, indicating strong detail preservation without introducing artifacts. In particular, we obtain an LPIPS of 0.053 and a PSNR of 25.93, while Champ ranks second in LPIPS and GUAVA ranks second in SSIM and PSNR. We also achieve an SSIM of 0.938 and an L1 error of 4.06, compared to 9.59 for AnimateAnyone (second best). 

For pose accuracy, \cref{tab:akd_csim_comparison} reports Average Keypoint Distance (AKD) for face, hands, and torso. Our method achieves the lowest AKD on hands and torso, and remains competitive on face AKD (0.18), close to GUAVA (0.15). Finally, we obtain the highest CSIM score of 0.85, demonstrating strong facial identity preservation.
We evaluate temporal consistency using the Temporal Jittering Error (TJE)~\cite{javanmardi2026talkingpose} over 1k-frame intervals. As shown in \cref{fig:TJE}, diffusion-based methods exhibit higher jitter due to their stochastic nature, with performance degrading as sequence length increases. While recent approaches incorporate motion modules that can produce stable results for short clips, they still face limitations in GPU memory requirements and struggle to maintain temporal consistency over longer sequences. In contrast, graphics-based approaches, including \textsc{GUAVA} and ours, achieve lower TJE and maintain stable temporal dynamics across extended durations.

\begin{table}[!t]
\caption{\textbf{Quantitative comparison on our captured multi-view human upper-body dataset.} Best results are in \textbf{bold} and second best are \underline{underlined}.}
\centering
\small
\setlength{\tabcolsep}{4pt}
\resizebox{0.9\linewidth}{!}{%
\begin{tabular}{@{}lrrrr@{}}
\toprule
\textbf{Method}
 & \multicolumn{1}{c}{\scriptsize L1 $\downarrow$}
 & \multicolumn{1}{c}{\scriptsize SSIM $\uparrow$}
 & \multicolumn{1}{c}{\scriptsize PSNR $\uparrow$}
 & \multicolumn{1}{c}{\scriptsize LPIPS $\downarrow$}\\
\midrule
MagicAnimate \cite{xu2024magicanimate}
  & 17.61 & 0.823 & 14.90 & 0.171 \\
AnimateAnyone \cite{hu2024animate}
  & \underline{9.59} & 0.859 & 19.62 & 0.132 \\
Champ \cite{zhu2024champ}
  & 11.07 & 0.890 & 19.36 & \underline{0.102} \\
GUAVA \cite{Zhang_2025_ICCV}
  & 15.24 & \underline{0.897} & \underline{22.33} & 0.118 \\
\midrule
Ours
  & \textbf{4.06} & \textbf{0.938} & \textbf{25.93} & \textbf{0.053} \\
\bottomrule
\end{tabular}%
}
\vspace{-16pt}
\label{tab:overall_comparison}
\end{table}

\begin{table}[!t]
\caption{\textbf{Quantitative comparison using keypoint and identity metrics.}
AKD denotes Average Keypoint Distance for face (F), hands (H), and Torso (T).}
\centering
\small
\setlength{\tabcolsep}{4pt}
\resizebox{0.9\linewidth}{!}{%
\begin{tabular}{@{}lrrrr@{}}
\toprule
\textbf{Method}
 & \multicolumn{1}{c}{\scriptsize AKD\_F $\downarrow$}
 & \multicolumn{1}{c}{\scriptsize AKD\_H $\downarrow$}
 & \multicolumn{1}{c}{\scriptsize AKD\_T $\downarrow$}
 & \multicolumn{1}{c}{\scriptsize CSIM $\uparrow$} \\
\midrule
MagicAnimate \cite{xu2024magicanimate}
  & 0.82 & 2.81 & 4.92 & 0.35 \\
AnimateAnyone \cite{hu2024animate}
  & 0.37 & 1.98 & 6.14 & 0.54 \\
Champ \cite{zhu2024champ}
  & 0.41 & 2.27 & 5.14 & 0.51 \\
GUAVA \cite{Zhang_2025_ICCV}
  & \textbf{0.15} & \underline{0.76} & \underline{1.39} & \underline{0.82} \\
\midrule
Ours
  & \underline{0.18} & \textbf{0.71} & \textbf{1.37} & \textbf{0.85} \\
\bottomrule
\end{tabular}%
}
\vspace{-13pt}
\label{tab:akd_csim_comparison}
\end{table}

\begin{table*}[!h]
\caption{\textbf{Ablation study on Novel-View synthesis and Self-Reenactment.} ``Upp. Body FT'' denotes fine-tuning upper-body mesh parameters. Best results are shown in \textbf{bold}.}
\centering
\small
\setlength{\tabcolsep}{6pt}
\resizebox{0.9\linewidth}{!}{%
\begin{tabular}{l|cccc|cccc}
\toprule
 & \multicolumn{4}{c|}{\textbf{Novel-View}} 
 & \multicolumn{4}{c}{\textbf{Self-Reenactment}} \\
\cmidrule(lr){2-5} \cmidrule(lr){6-9}
\textbf{Method}
 & L1 $\downarrow$ & PSNR $\uparrow$ & SSIM $\uparrow$ & LPIPS $\downarrow$
 & L1 $\downarrow$ & PSNR $\uparrow$ & SSIM $\uparrow$ & LPIPS $\downarrow$ \\
\midrule
\textbf{Ours}
 & \textbf{2.46} & \textbf{24.94} & \textbf{0.958} & \textbf{0.074}
 & \textbf{2.71} & 24.18 & 0.953 & \textbf{0.075} \\

w/o FLAME \& MANO
 & 3.08 & 24.15 & 0.947 & 0.083
 & 3.08 & 23.35 & 0.943 & 0.082 \\

w/o Upp. \textsc{Body} FT
 & 2.67 & 24.10 & 0.953 & 0.080
 & 2.74 & \textbf{24.24} & \textbf{0.954} & 0.077 \\

w/o LPIPS
 & 4.90 & 24.25 & 0.943 & 0.085
 & 4.54 & 23.17 & 0.953 & 0.098 \\
\bottomrule
\end{tabular}%
}
\label{tab:ablation_subject004}
\end{table*}

\subsubsection{Qualitative Results}
\label{subsubsec:qualitative_results}
\textbf{Self-reenactment.} As shown in \cref{fig:qualitative}, MagicAnimate, AnimateAnyone, and Champ often introduce background artifacts and fail to reproduce fine-grained hand gestures, while also exhibiting weaker identity preservation. In contrast, both GUAVA and our method generate sharper renderings with accurate pose and expression. However, GUAVA frequently loses facial identity details, whereas our approach preserves identity more faithfully and produces more photorealistic results. We further observe that our method remains robust under challenging hand articulations involving severe finger self-occlusions and fine-grained finger interactions. As illustrated in \cref{fig:handy}, competing methods often produce distorted finger configurations or fail to preserve the intended gesture, whereas our approach accurately reconstructs the hand shape while maintaining appearance consistency.

\noindent \textbf{Novel-view synthesis.} Since novel-view rendering is essential for 3D applications, we additionally compare against GUAVA, as the other baselines do not support view extrapolation. In \cref{fig:novel}, we fix pitch and roll and vary the yaw angle to $+5^\circ$ and $+20^\circ$. Our method maintains consistent geometry and high-quality appearance under viewpoint changes, while GUAVA degrades as the viewpoint moves further away from the training cameras.

\noindent \textbf{Cross-identity animation.} We further evaluate cross-identity animation by transferring motion from one subject to another. As shown in \cref{fig:cross}, our method preserves the driving gestures while maintaining the target identity.
\vspace{-8pt}
\subsection{Ablation Studies}
\label{subsec:ablation}

We conduct ablation studies to quantify the impact of key components in our avatar synthesis pipeline. \cref{tab:ablation_subject004} reports results on a held-out subject under both self-reenactment and novel-view synthesis settings, allowing us to evaluate both motion reenactment quality and cross-view generalization.

\subsubsection{Without FLAME and MANO.}
We remove FLAME and MANO and optimize only the upper-body parameters (body pose, global rotation, and translation). This leads to a clear drop across all metrics in both settings. 
Qualitatively, the reconstructed avatars exhibit less accurate facial expressions and degraded hand articulation, which noticeably reduces realism and identity consistency. These results confirm that explicit face and hand modeling is crucial for high-fidelity avatar reconstruction and reenactment.
\vspace{-8pt}
\subsubsection{Without upper-body parametric fine-tuning.}
We disable upper-body parametric fine-tuning during training and rely solely on the pre-fitted parameters obtained from multi-view keypoints and FLAME/MANO fits. Performance decreases in both settings, indicating that jointly refining the parametric model together with Gaussian appearance is important for accurate reconstruction.

\subsubsection{Without LPIPS loss.}
Finally, we remove the LPIPS term from the RGB reconstruction objective. This leads to lower perceptual quality, showing that LPIPS provides complementary supervision beyond pixel-wise losses and improves overall visual fidelity.


\subsection{Ethical Considerations and Potential Misuse.}
Our method enables high-fidelity human avatar reconstruction and cross-identity animation, which can be misused for impersonation, deceptive media generation, or non-consensual content. We emphasize that such uses are unethical and strongly discourage them. The dataset used in this work contains identifiable facial and motion data; all participants provided informed consent for data capture and research use, following applicable institutional guidelines. We recommend that future releases of data or models include appropriate safeguards, such as usage restrictions, consent-based data sharing, and disclosure or watermarking mechanisms, to mitigate potential misuse and promote responsible deployment.

\subsection{Limitations}
\label{subsec:limitations}
Despite achieving consistent avatar reconstruction and reenactment, our approach still relies on a parametric model. While this representation is stable and easy to tune, it is difficult to obtain. 
In addition, full 360$^\circ$ rendering remains challenging with the current dataset and pipeline, as we restrict training and evaluation to frontal views where most appearance cues are visible. 
Moreover, our model does not explicitly model secondary motion (e.g., clothing and accessories deformation), which can reduce realism under fast or complex movements.

\section{Conclusion}
\label{sec:conclusion}
We presented \textbf{MVFGA}, a multi-view framework for high-fidelity upper-body avatars with fine-grained facial expressions and articulated hand motion.
Our key idea is to build a unified upper-body avatar representation by explicitly integrating detailed face and hand parameterizations into an upper-body parametric model, and coupling it with a deformable 3D Gaussian appearance field for photorealistic novel-view rendering.
We evaluate MVFGA on our captured multi-view upper-body dataset under self-reenactment, novel-view synthesis, and cross-identity animation. 
MVFGA consistently outperforms strong generative and graphics-based baselines across image-quality metrics, producing sharper renderings with better identity preservation and more accurate facial and hand motion. 
Keypoint-based evaluation shows competitive pose accuracy, and ablations verify the importance of explicit FLAME and MANO integration, upper-body parametric fine-tuning during Gaussian optimization, and LPIPS supervision. 
We also introduce \textbf{MVFGA-MoCap}, a synchronized, calibrated multi-view dataset with fitted parametric models to support future research on upper-body avatar reconstruction and animation.
We hope this research will support continued progress on high-fidelity upper-body avatars and encourage further exploration of multi-view datasets and representation.
\section{Acknowledgments}
This work was partially funded by the Horizon Europe programme under the project IRIS-XR, Grant Agreement No. 101298672.

\printbibliography 

\clearpage
\appendix
\section{MVFGA-MoCap Dataset: Supplementary Material}
\label{sec:supp_dataset}

\setcounter{figure}{0}
\renewcommand{\thefigure}{S\arabic{figure}}
\renewcommand{\theHfigure}{supp.\arabic{figure}}
\setcounter{table}{0}
\renewcommand{\thetable}{S\arabic{table}}
\renewcommand{\theHtable}{supp.\arabic{table}}
\setcounter{equation}{0}
\renewcommand{\theequation}{S\arabic{equation}}
\renewcommand{\theHequation}{supp.\arabic{equation}}
\raggedbottom

This supplementary material provides detailed documentation of the \textbf{MVFGA-MoCap} dataset, a custom multi-view motion capture dataset collected to support high-fidelity upper-body avatar reconstruction and animation. The dataset is designed to capture fine-grained facial expressions, hand gestures, and torso motion under controlled multi-view conditions.

\subsection{Multi-View Capture Setup}
\label{subsec:capture_setup}

Data acquisition is performed in a dedicated motion capture studio using a rigid multi-camera rig. The final setup consists of \textbf{17 synchronized DSLR cameras}. Among these, \textbf{15 cameras} are mounted on a structured frontal rig covering approximately $150^\circ$ around the subject, while \textbf{two additional cameras} are positioned at rear-side viewpoints to improve triangulation and depth estimation accuracy.

As shown in \cref{Fig:config}, all cameras are Nikon D3200 DSLRs capturing video at a resolution of $1920 \times 1080$ pixels and 25 FPS. To ensure uniform illumination and minimize shadows, flicker-free studio lights are rigidly mounted on the same aluminum profiles as the cameras in a triangular configuration. This design ensures geometric stability, lighting consistency, and repeatability across recording sessions.

\begin{center}
  \centering
  \includegraphics[width=0.9\columnwidth]{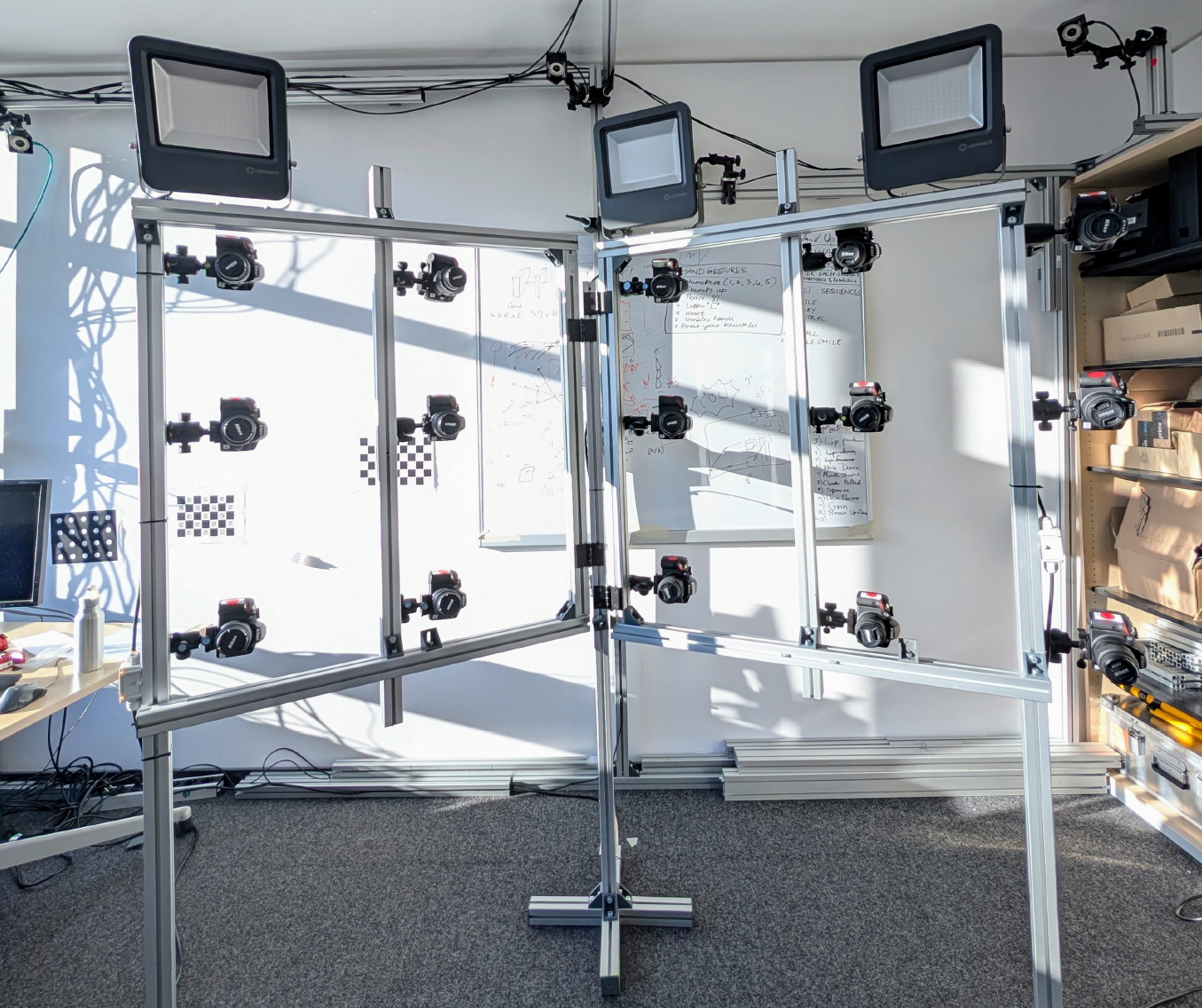}
\vspace{-5pt}
\captionof{figure}{Multi-view camera setup with lights based on our design. This figure shows cameras
and lights mounted on the aluminium profile with a $150^\circ$ field of view.}
  \label{Fig:config}
\end{center}

\subsection{Participant Consent and Data Collection Ethics}
All participants were informed about the purpose of the data collection and the intended research use of the recorded material. Prior to recording, each participant was provided with a written consent form describing the capture procedure, the use of the data for research and publication purposes, and their right to withdraw from the study. Participation was voluntary, and all recordings were conducted only after obtaining informed consent from the participants.

\subsection{Camera Calibration}
\label{subsec:calibration}

All cameras are calibrated using a checkerboard-based calibration pipeline implemented in OpenCV \cite{opencv_library}. The objective is to estimate both the intrinsic parameters of each camera (focal length, principal point, skew, and lens distortion) and the extrinsic parameters (rotation and translation) that describe the camera pose with respect to the world coordinate system.

\subsubsection{Pinhole camera model.}
Let a 3D point in the world coordinate system be denoted by
$\mathbf{X}_w = [X_w,\, Y_w,\, Z_w]^\top$.
Using the camera extrinsic parameters, the point is first transformed from world coordinates into camera coordinates:
\begin{equation}
\label{eq:world_to_camera}
\begin{bmatrix}
\mathbf{X}_c \\
1
\end{bmatrix}
=
\begin{bmatrix}
\mathbf{R} & \mathbf{t} \\
\mathbf{0}_{1\times 3} & 1
\end{bmatrix}
\begin{bmatrix}
\mathbf{X}_w \\
1
\end{bmatrix},
\end{equation}
where $\mathbf{R}\in\mathbb{R}^{3\times 3}$ is the rotation matrix and $\mathbf{t}\in\mathbb{R}^{3}$ is the translation vector.

The 3D camera-space point $\mathbf{X}_c = [X_c,\,Y_c,\,Z_c]^\top$ is then projected onto the image plane using the intrinsic calibration matrix $\mathbf{K}$:
\begin{equation}
\label{eq:projection_intrinsic}
\lambda
\begin{bmatrix}
u\\
v\\
1
\end{bmatrix}
=
\mathbf{K}
\begin{bmatrix}
\mathbf{I}_3 & \mathbf{0}_3
\end{bmatrix}
\begin{bmatrix}
\mathbf{X}_c\\
1
\end{bmatrix},
\end{equation}
where $(u,v)$ are the pixel coordinates, $\lambda$ is a projective scale factor, and $\mathbf{I}_3$ is the $3\times 3$ identity matrix.

Combining \eqref{eq:world_to_camera} and \eqref{eq:projection_intrinsic} yields the standard full projection equation:
\begin{equation}
\label{eq:full_projection}
\lambda
\begin{bmatrix}
u\\
v\\
1
\end{bmatrix}
=
\mathbf{K}
\begin{bmatrix}
\mathbf{R} & \mathbf{t}
\end{bmatrix}
\begin{bmatrix}
X_w\\
Y_w\\
Z_w\\
1
\end{bmatrix}
=
\mathbf{P}
\begin{bmatrix}
X_w\\
Y_w\\
Z_w\\
1
\end{bmatrix},
\end{equation}
where $\mathbf{P}\in\mathbb{R}^{3\times 4}$ is the camera projection matrix.

\subsubsection{Intrinsic matrix.}
The intrinsic matrix $\mathbf{K}$ is modeled as an upper-triangular matrix:
\begin{equation}
\label{eq:K_matrix}
\mathbf{K}
=
\begin{bmatrix}
f_x & s   & c_x\\
0   & f_y & c_y\\
0   & 0   & 1
\end{bmatrix},
\end{equation}
where $(f_x,f_y)$ are the focal lengths in pixel units, $(c_x,c_y)$ is the principal point, and $s$ is the skew parameter.

\subsubsection{Checkerboard-based calibration.}
Calibration images are acquired by capturing multiple views of a planar checkerboard pattern from different angles and distances. Since the 3D geometry of the checkerboard corners is known \textit{a priori}, the calibration procedure detects the corresponding 2D corner locations in each image (e.g., using Harris corner detection followed by sub-pixel refinement). The set of 3D--2D correspondences is then used to estimate $\mathbf{K}$, lens distortion coefficients, and per-image extrinsic parameters $(\mathbf{R},\mathbf{t})$ by minimizing the re-projection error between observed and predicted corner locations.

\subsubsection{Camera pose estimation (PnP).}
Given a set of $N$ correspondences $\{(\mathbf{X}_w^{(i)},\, \mathbf{x}^{(i)})\}_{i=1}^N$, where $\mathbf{x}^{(i)}=[u^{(i)},v^{(i)}]^\top$ are image measurements, the camera pose is computed using a Perspective-\textit{n}-Point (PnP) formulation. In practice, OpenCV solves for $(\mathbf{R},\mathbf{t})$ using iterative Levenberg--Marquardt optimization, minimizing:
\begin{equation}
\label{eq:reprojection_error}
\min_{\mathbf{R},\mathbf{t}} \;
\sum_{i=1}^{N}
\left\|
\mathbf{x}^{(i)}
-
\pi\!\left(\mathbf{K},\mathbf{R},\mathbf{t},\mathbf{X}_w^{(i)}\right)
\right\|_2^2,
\end{equation}
where $\pi(\cdot)$ denotes the (distortion-aware) projection function from 3D world coordinates to 2D pixel coordinates.

The resulting intrinsic and extrinsic parameters define the projection matrix $\mathbf{P}$ for each camera, enabling accurate mapping between 3D world coordinates and 2D image pixels. This calibration is essential for consistent multi-view alignment and reliable triangulation of 3D keypoints.

\subsection{Upper Body MoCap Dataset}
\label{subsec:upper_body_dataset}

Our research focuses on recording the complete upper body, including the face, hands, and torso, in order to create a high-quality multi-view motion capture dataset that preserves subtle facial expressions, body poses, and hand gestures. This level of detail is particularly valuable for applications such as avatar-based weather forecasting, instructional and educational content creation, virtual conferencing, medical imaging, and immersive Augmented and Virtual Reality (AR/VR) experiences.

\subsubsection{Dataset Scenario}
\label{subsubsec:dataset_scenario}

The dataset is captured using a multi-camera setup arranged with an approximate $150^{\circ}$ angular coverage to ensure broad visibility of the subject and maximize temporal and spatial alignment across views. During the acquisition process, a manual synchronization protocol is employed to guarantee frame-level consistency between the recorded camera streams.

Each recording session begins with the participant performing a distinct hand clap gesture, which is clearly visible across all camera views and serves as a temporal reference point for synchronization. After the initial alignment cue, the subject performs a predefined sequence of facial expressions and hand gestures designed to capture a wide range of upper-body motion variations. This standardized protocol is consistently followed by all participants, enabling controlled data collection suitable for downstream tasks such as self-reenactment and cross-reenactment of digital avatars. The session concludes with a second hand clap, providing an additional temporal anchor for verification. Finally, all recorded sequences are manually synchronized and temporally trimmed to ensure precise frame-level consistency across the full set of video streams.

\subsubsection{Statistics}
\label{subsubsec:dataset_statistics}

The participant distribution in the dataset is designed to be relatively balanced across gender and age. In total, the dataset includes $15$ participants, consisting of $8$ males and $7$ females, with ages ranging from $24$ to $32$ years. This distribution provides diversity in terms of demographic attributes, which contributes to the robustness and generalizability of models trained on the dataset.

\subsubsection{Hand Gestures}
\label{subsubsec:hand_gestures}

Table~\ref{tab:hand_gestures} summarizes the set of hand gestures performed by the participants during data acquisition. All gestures are executed using both hands and captured from multiple camera views. For each frame, 2D keypoints are extracted from all available views and triangulated to obtain 3D keypoints. The selected gesture set was designed to cover a diverse range of hand articulations that are particularly useful for avatar animation and hand motion modeling.

\begin{table}[t]
    \centering
    \caption{Collection of hand gestures performed by participants during dataset recording.}
    \label{tab:hand_gestures}
    \begin{tabular}{l c}
        \hline
        \textbf{Gestures} & \textbf{\# of Images} \\
        \hline
        Clapping & 2K \\
        Conversational & 560K \\
        Heart symbol & 3K \\
        Letter ``L'' & 3.5K \\
        Number 1,2,3,4 and 5 & 9.5K \\
        Pushing finger tips & 18K \\
        Thumbs-up & 5K \\
        Two hands counting extension & 10K \\
        Universal Peace sign & 4K \\
        \hline
    \end{tabular}
\end{table}

\subsubsection{Facial Expressions}
\label{subsubsec:facial_expressions}

In addition to hand motion, the dataset contains a diverse set of facial expressions captured from $15$ front-facing cameras positioned at different angles. The facial expressions are parameterized using the FLAME model, and the resulting sequences are divided into training and test splits to ensure a balanced distribution of seen and unseen expressions. This design supports robust evaluation of facial motion generalization across subjects and expression categories.

Table~\ref{tab:facial_expressions} reports the facial expression classes included in the dataset along with the corresponding number of images for each category.

\begin{table}[t]
\centering
\captionof{figure}{Facial expressions captured in our dataset, each recorded across multiple camera views.}
\label{tab:facial_expressions}
\small
\setlength{\tabcolsep}{6pt}
\renewcommand{\arraystretch}{1.1}

\begin{tabular}{l r}
\toprule
\textbf{Expression} & \textbf{\# of Images} \\
\midrule
Angry & 1.5K \\
Cheeks puffed & 2K \\
Chin raiser & 1.2K \\
Eyes closed/open & 1.8K \\
Grin & 1K \\
Jaw L/R & 3K \\
Lip puckerer & 1.8K \\
Mouth open & 3.5K \\
Mouth stretched & 3.3K \\
Nose down & 1.2K \\
Neutral & 50K \\
Sad & 3.5K \\
Smile & 8K \\
Squeeze & 2.6K \\
\bottomrule
\end{tabular}
\end{table}


\subsection{Dataset Post-Processing}
\label{subsec:post_processing}

This section provides an overview of the post-processing pipeline applied to the captured multi-view recordings. First, temporally synchronized video streams are processed to extract 2D human keypoints using MediaPipe \cite{lugaresi2019mediapipe}. Next, parametric human body models are fitted to these observations, including FLAME for facial modeling, MANO \cite{MANO:SIGGRAPHASIA:2017} for hand modeling, and SMPL-X \cite{SMPL-X:2019} for full-body fitting. In parallel, body-part segmentation and background matting are performed to isolate the upper-body region and reduce background artifacts. Together, these steps ensure that the resulting data is clean, well-aligned, and suitable for reliable model fitting.

\subsubsection{Body Part Segmentation}
\label{subsubsec:body_part_segmentation}

To obtain body-part segmentation masks for each frame, we feed individual images from the multi-camera setup into the \textit{Sapiens} human parsing model~\cite{khirodkar2024sapiens}. For each frame, the predicted semantic labels are used to retain only the classes corresponding to the upper-body region, while all remaining pixels (including the background and lower-body parts) are set to white. This preprocessing maintains visual consistency across views and ensures that the downstream reconstruction pipeline focuses on the foreground subject.

In addition to improving the visual quality of the training data, the segmentation masks are also used as auxiliary constraints during model fitting. In particular, they help reduce the mask loss and improve the accuracy of parametric model optimization compared to using keypoints alone. \cref{Fig:seg} illustrates an example output of the human parsing model, where different body parts are assigned unique colors.

\begin{center}
  \centering
  \includegraphics[width=0.9\columnwidth]{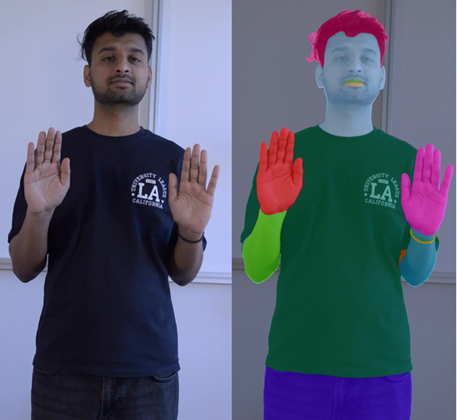}
\vspace{-5pt}
\captionof{figure}{Body part segmentation output from \cite{khirodkar2024sapiens} including the ground truth image for 2
subjects from the center camera.}
  \label{Fig:seg}
\end{center}

\subsubsection{Background Matting}
\label{subsubsec:background_matting}

To further isolate the subject from the background, we apply a background matting technique that removes the background region from each frame. We utilize a bilateral reference network based on Dichotomous Image Segmentation (DIS)~\cite{Qin_2020_PR}, which consists of two main components: a \emph{localization module} and a \emph{referencing module}. Using the body-part segmentation results from Section~\ref{subsubsec:body_part_segmentation}, we select the upper-body region and mask out the remaining pixels, setting them to white.
\begin{center}
  \centering
  \includegraphics[width=0.9\columnwidth]{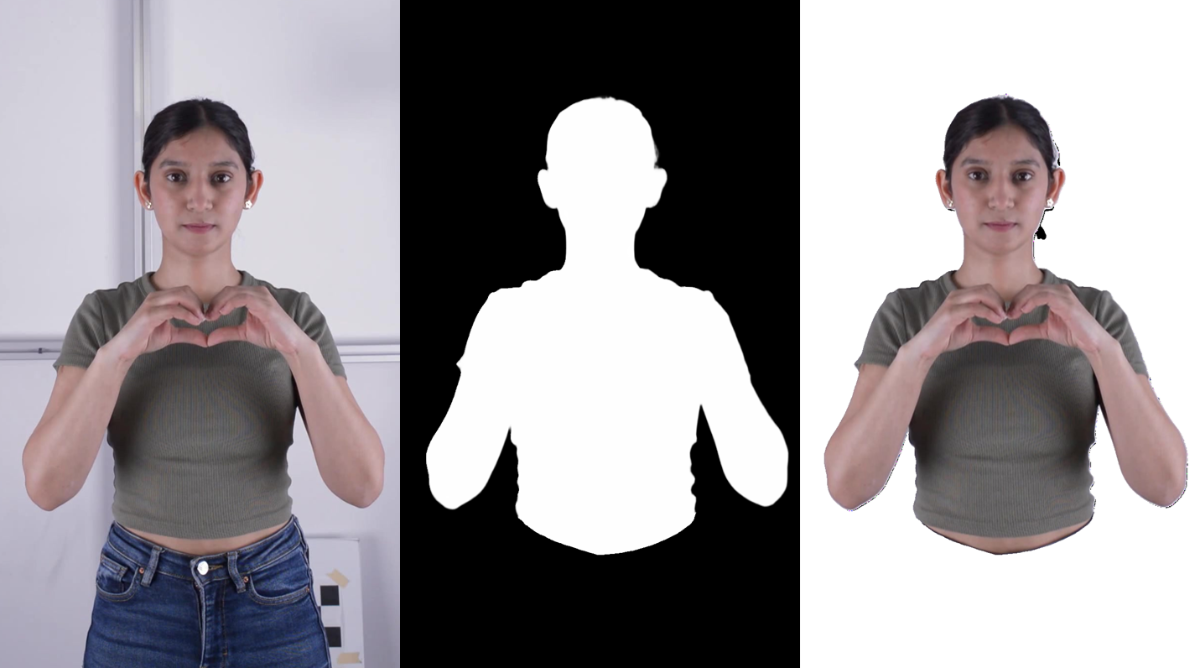}
\vspace{-5pt}
\captionof{figure}{Output from BiRefNet \cite{zheng2024birefnet}. Starting from left to right, we have input image
(with background), input image (w/o background and lower body) and alpha
map for the modified input image (center).}
  \label{Fig:fore}
\end{center}
\cref{Fig:fore} shows qualitative examples of the matting results, including the original input image, the updated image with a white background and removed lower-body region, and the predicted alpha matte. This approach reduces noisy reconstruction artifacts, minimizes background inconsistencies across different viewpoints, and improves model generalization, enabling the resulting avatars to be more easily transferred across devices and environments.

\subsubsection{Pose Estimation}
\label{subsubsec:pose_estimation}

For pose estimation, we provide individual frames from all camera views as input to the MediaPipe pose estimation module. The extracted 2D keypoints are then used as observations for parametric model fitting (FLAME, MANO, and SMPL-X). The optimization pipeline minimizes the discrepancy between the projected 3D joints and the detected 2D keypoints. 

\subsubsection{FLAME Parameters}
\label{subsubsec:flame_parameters}

To track facial motion and expressions, we employ the FLAME face model~\cite{FLAME:SiggraphAsia2017}, which represents facial geometry using parameters for identity shape, expression, eye pose, and jaw pose. To fit FLAME to the captured image sequences, we adopt the Metrical Tracker framework~\cite{MICA:ECCV2022}, an optimization-based face tracking approach originally designed for monocular RGB sequences. The framework models non-rigid deformations using linear expression bases and linear blend skinning, while facial appearance is represented using an albedo model under Lambertian reflectance assumptions. Illumination is approximated using spherical harmonics, and the fitting process follows an analysis-by-synthesis strategy.
\begin{center}
  \centering
  \includegraphics[width=0.9\columnwidth]{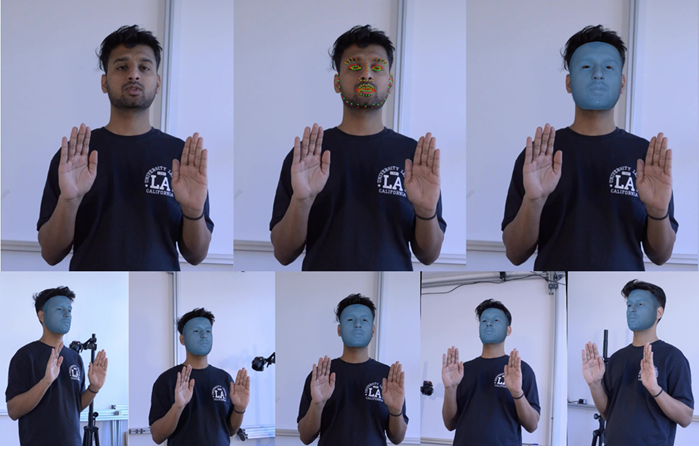}
\vspace{-5pt}
\captionof{figure}{Output visualization from Metrical Tracker. Starting from top to bottom, we have
multi-view input images, dense photometric term, 2D landmark keypoints con-
taining green and red points representing ground truth and predicted keypoints
respectively and finally, face mask of the FLAME model reprojected onto the input
images.}
  \label{Fig:flame}
\end{center}
The identity shape parameters, denoted by $\boldsymbol{\beta}$, are predicted using the MICA shape regressor~\cite{MICA:ECCV2022}. In our implementation, we extend the original Metrical Tracker framework to incorporate multi-view information by jointly leveraging 2D keypoints extracted from all 15 front-facing cameras. This multi-view extension provides stronger geometric constraints and improves robustness under challenging head poses and complex expressions.

\textbf{Initialization.}
The albedo parameters and spherical harmonics lighting coefficients are initialized by optimizing the first frame of each input sequence using the following energy function:
\begin{equation}
\label{eq:flame_energy}
E(\phi) = w_{\text{dense}} E_{\text{dense}}(\phi)
        + w_{\text{lmk}} E_{\text{lmk}}(\phi)
        + w_{\text{reg}} E_{\text{reg}}(\phi),
\end{equation}
where $\phi$ denotes the parameters being optimized. Here, $E_{\text{lmk}}$ is the 2D landmark re-projection error, $E_{\text{reg}}$ is a regularization term that prevents unrealistic deformations, and $E_{\text{dense}}$ enforces dense photometric consistency using an $\ell_1$-norm. All optimization steps are performed using the Adam optimizer in PyTorch. Visualization is shown in \cref{Fig:flame}.

\subsubsection{MANO Parameters}
\label{subsubsec:mano_parameters}

To accurately reconstruct detailed hand gestures, we process the hand regions from each input view independently. We employ the Hand Mesh Recovery (HaMeR) model~\cite{pavlakos2024reconstructing}, which leverages the MANO parametric hand model~\cite{MANO:SIGGRAPHASIA:2017}. As shown in \cref{Fig:mano}, MANO represents hand geometry using pose parameters $\boldsymbol{\theta}\in\mathbb{R}^{48}$ and shape parameters $\boldsymbol{\beta}\in\mathbb{R}^{10}$, and outputs a 3D hand mesh with 778 vertices along with 3D joint locations:
\begin{equation}
\mathbf{X}\in\mathbb{R}^{K\times 3}, \qquad K=21.
\end{equation}

\begin{center}
  \centering
  \includegraphics[width=0.9\columnwidth]{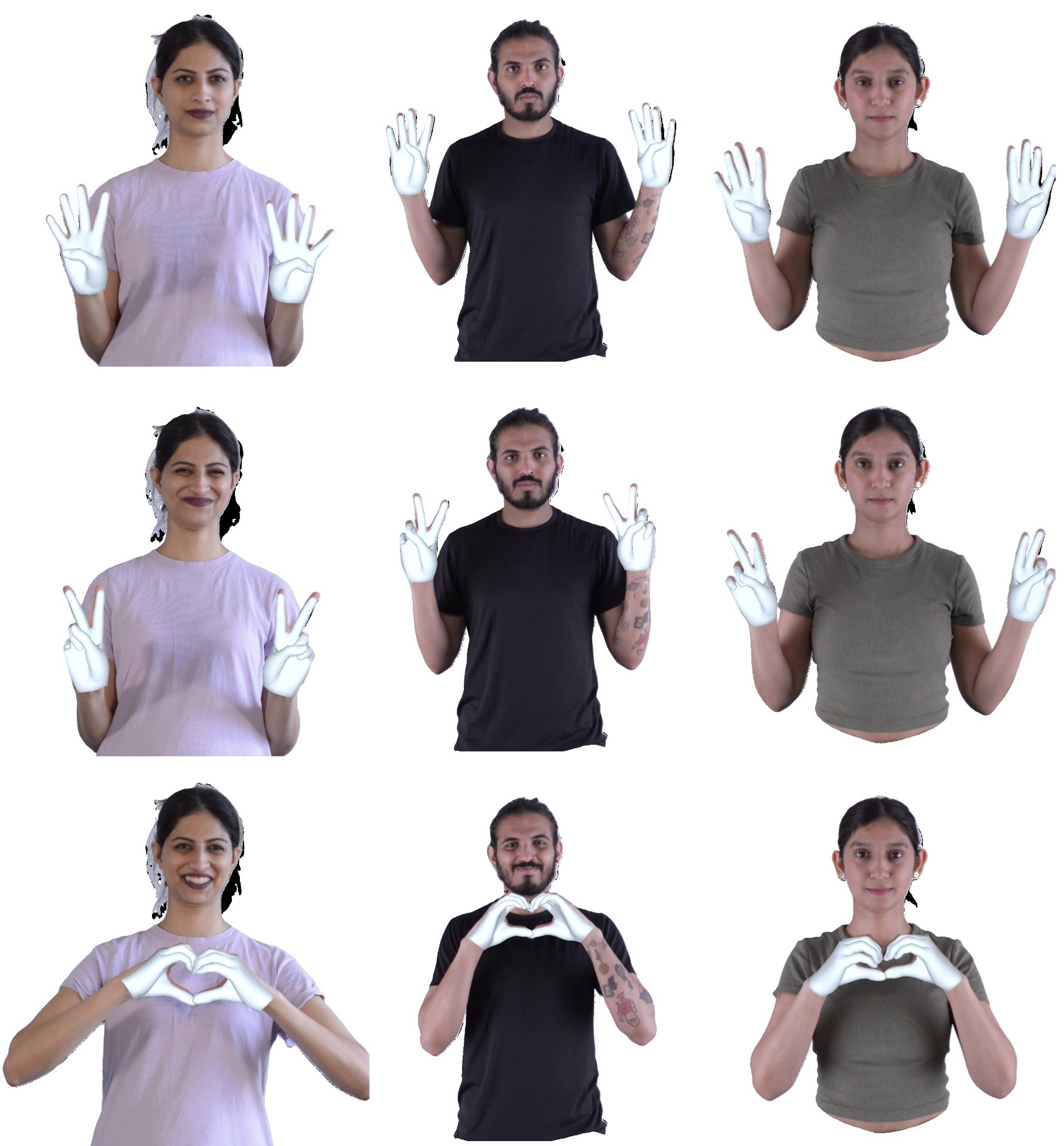}
\vspace{-5pt}
\captionof{figure}{MANO prediction for three different hand gestures (number 4, heart symbol,
universal peace sign) performed by 3 (1 male/2 female) different participants
during the dataset capture.}
\vspace{+15pt}
  \label{Fig:mano}
\end{center}

In addition to estimating MANO parameters from a single RGB image, HaMeR also predicts camera parameters $\boldsymbol{\pi}$, enabling projection of the reconstructed mesh onto the image plane. The model is based on a transformer architecture \cite{50650} using a Vision Transformer (ViT-H) backbone to encode the input image into visual tokens, followed by a transformer decoder head that outputs the parameter set:
\begin{equation}
\boldsymbol{\Theta} = (\boldsymbol{\theta}, \boldsymbol{\beta}, \boldsymbol{\pi}).
\end{equation}

In our implementation, we run the pretrained HaMeR model on all front-facing camera views. To improve geometric precision, we adapt the pipeline to incorporate calibrated camera intrinsics, allowing us to undistort the input images prior to inference. The predicted hand pose parameters and joint coordinates are then integrated into the SMPL-X fitting pipeline, improving the realism and fidelity of hand reconstruction.

\clearpage
\balance
\subsubsection{SMPL-X Parameters}
\label{subsubsec:smplx_parameters}
\begin{center}
  \centering
  \includegraphics[width=0.9\columnwidth]{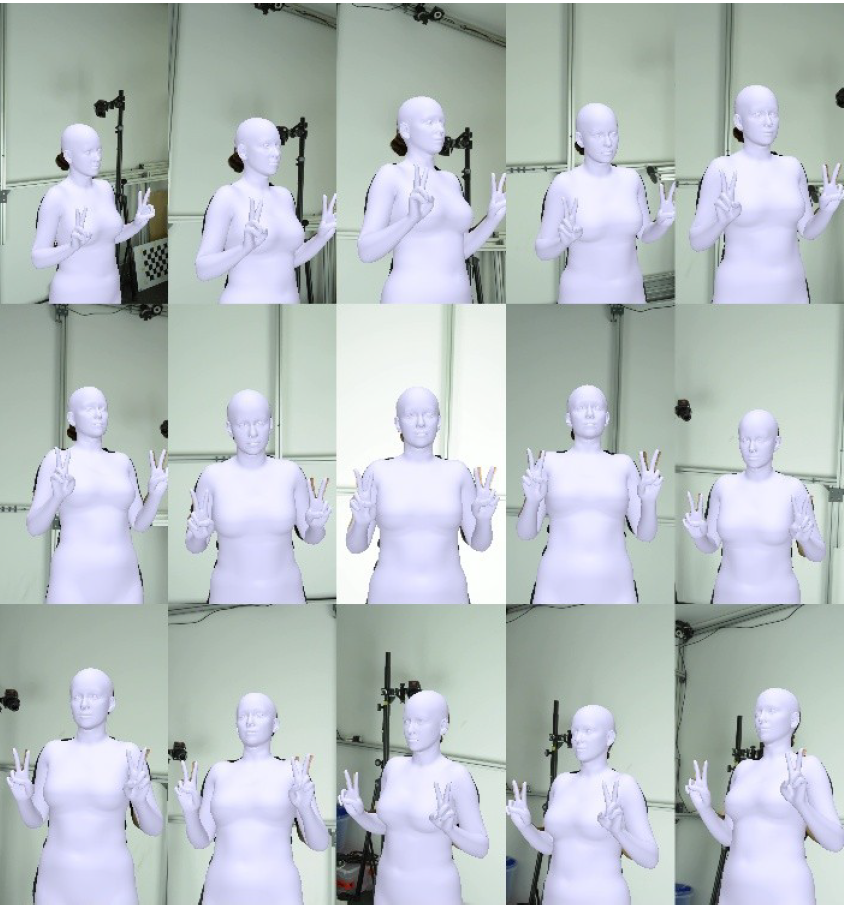}
\vspace{-5pt}
\captionof{figure}{SMPL-X fitting from EasyMocap for one timestep of one of the subjects. This
shows the fitting from all front-facing cameras.}
  \label{Fig:mesh}
\end{center}
Following the estimation of FLAME and MANO parameters, we integrate them into a unified SMPL-X representation. While SMPL-X~\cite{SMPL-X:2019} provides parameters for face, hands, and body, its default configuration does not capture fine-grained facial expressions and hand articulation with the same fidelity as FLAME and MANO. To address this, we adopt a parameter merging strategy where refined facial and hand parameters from FLAME and MANO are injected into SMPL-X while retaining the global body pose, orientation, and translation.

In this study, we modify the single-person SMPL-X fitting code from EasyMocap~\cite{dong2021fast} to accept these additional parameter inputs. The fitting pipeline begins by extracting 2D keypoints using MediaPipe's holistic model. These keypoints are reformatted to match the OpenPose \cite{8765346} convention with 25 body joints. Using triangulation across multiple views, we estimate 3D joint positions and filter unreliable joints based on confidence scores and re-projection errors.

The SMPL-X optimization aims to estimate the parameter set:
\begin{equation}
\boldsymbol{\theta} = \{\boldsymbol{\beta},\, \boldsymbol{\theta}_{\text{body}},\, \boldsymbol{\theta}_{\text{hands}},\, \boldsymbol{\theta}_{\text{face}},\, \mathbf{R},\, \mathbf{T}\},
\end{equation}
where $\boldsymbol{\beta}$ denotes body shape parameters, $\boldsymbol{\theta}_{\text{body}}$, $\boldsymbol{\theta}_{\text{hands}}$, and $\boldsymbol{\theta}_{\text{face}}$ represent pose parameters for the body, hands, and face, and $(\mathbf{R},\mathbf{T})$ correspond to global rotation and translation.

To estimate the optimal parameters, we use the Limited-memory Broyden--Fletcher--Goldfarb--Shanno (L-BFGS) optimization algorithm~\cite{liu1989limited}. The objective is to minimize the following loss:
\begin{equation}
\label{eq:smplx_loss}
\begin{aligned}
\min_{\boldsymbol{\theta}} \mathcal{L}(\boldsymbol{\theta})
&=
\lambda_{2D}\mathcal{L}_{2D}
+\lambda_{3D}\mathcal{L}_{3D}
+\lambda_{\text{limb}}\mathcal{L}_{\text{limb}}
\\
&\quad+
\lambda_{\text{smooth}}\mathcal{L}_{\text{smooth}}
+\lambda_{\text{reg}}\mathcal{L}_{\text{reg}}.
\end{aligned}
\end{equation}

The optimization is performed in multiple stages to improve convergence and robustness. First, an initial pose estimation step aligns the global orientation and translation using pelvis-based priors and 2D keypoint alignment. Next, a full-body fitting stage refines pose and shape by minimizing joint re-projection errors in both 2D and 3D. Finally, when detailed facial and hand keypoints become available, they are incorporated to further refine the reconstruction. \cref{Fig:mesh} shows an example of SMPL-X fitting results, where the reconstructed mesh is rendered over the original multi-view images.

\subsection{Runtime Performance}

We further evaluate the runtime performance of our method to assess its suitability for real-time avatar animation. Our approach benefits from the efficiency of the 3D Gaussian Splatting representation, enabling fast upper-body avatar rendering while maintaining high visual fidelity. At inference time, the rendering speed mainly depends on the number of active Gaussians, which is controlled by our adaptive densification and pruning strategy. This provides a practical quality--speed tradeoff, where the representation can be adjusted to favor either higher visual detail or faster rendering.

On a single NVIDIA RTX 3090 GPU, our method achieves approximately 49 FPS, demonstrating real-time performance for upper-body face and gesture animation. This is significantly faster than diffusion-based animation methods such as Champ, which runs at 0.53 FPS, while remaining competitive with recent graphics-based avatar methods such as GUAVA, which achieves 52.21 FPS. These results show that our method combines the visual fidelity of an explicit 3D representation with the efficiency required for interactive avatar applications.

\end{document}